\documentclass[runningheads]{llncs}

\usepackage[T1]{fontenc}

\usepackage{graphicx,verbatim}
\usepackage{amsmath}
\usepackage{amssymb}
\usepackage{enumitem}
\usepackage{booktabs}
\usepackage{multirow}
\usepackage{adjustbox}
\usepackage{xcolor}
\usepackage{soul}
\definecolor{hlbluebg}{RGB}{217,217,255}
\definecolor{hlgreenbg}{RGB}{204,255,204}
\definecolor{hlredbg}{RGB}{255,204,204}
\newcommand{\hlblue}[1]{{\sethlcolor{hlbluebg}\hl{#1}}}
\newcommand{\hlgreen}[1]{{\sethlcolor{hlgreenbg}\hl{#1}}}
\newcommand{\hlred}[1]{{\sethlcolor{hlredbg}\hl{#1}}}
\usepackage{xurl}
\usepackage{float}

\begin{document}
\title{LLM-Bootstrapped Targeted Finding Guidance for Factual MLLM-based Medical Report Generation}
\titlerunning{Targeted Finding Guidance for Medical Report Generation}
\author{Cunyuan Yang\inst{1} \and
Dejuan Song\inst{2} \and
Xiaotao Pang\inst{3} \and
Qianqian Shen\inst{1} \and
Wenjie Nie \inst{1} \and
Yifan Huang \inst{1} \and
Lei Wu \inst{1} \and
Wei Han \inst{2} \and
Haishuai Wang \inst{1} \and
Jiajun Bu \inst{1}}
\authorrunning{Cunyuan Yang et al.}
\institute{Zhejiang University \and
The Second Affiliated Hospital Zhejiang University School of Medicine \and
Hangzhou Pu Jian Medical Technology Co., Ltd.\\
}

\maketitle

\begin{abstract}
    The automatic generation of medical reports utilizing Multimodal Large Language Models (MLLMs) frequently encounters challenges related to factual instability, which may manifest as the omission of findings or the incorporation of inaccurate information, thereby constraining their applicability in clinical settings. Current methodologies typically produce reports based directly on image features, which inherently lack a definitive factual basis. In response to this limitation, we introduce Fact-Flow, an innovative framework that separates the process of visual fact identification from the generation of reports. This is achieved by initially predicting clinical findings from the image, which subsequently directs the MLLM to produce a report that is factually precise. A pivotal advancement of our approach is a pipeline that leverages a Large Language Model (LLM) to autonomously create a dataset of labeled medical findings, effectively eliminating the need for expensive manual annotation. Extensive experimental evaluations conducted on two disease-focused medical datasets validate the efficacy of our method, demonstrating a significant enhancement in factual accuracy compared to state-of-the-art models, while concurrently preserving high standards of text quality.
\end{abstract}
\section{Introduction}
\label{sec:intro}
The automated generation of medical reports from diagnostic images is a critical task in computational medicine. Early approaches to Medical Report Generation (MRG) relied on encoder-decoder architectures, evolving from CNN-RNN combinations~\cite{jing2018automatic,wang2018tienet,xue2018multimodal} to Transformer-based models~\cite{chen2020generating,chen2021cross,nooralahzadeh2021progressive,liu2021exploring}. However, they struggled with capturing complex semantic relationships and generating factually precise narratives~\cite{miura2021improving}.

The recent advent of Multimodal Large Language Models (MLLMs)~\cite{liu2023visual} has introduced a new paradigm for MRG. MLLMs such as LLaVA-Med~\cite{li2023llava} have demonstrated immense potential, yet a significant challenge emerges when fine-tuned end-to-end: factual instability. These models are prone to hallucinating findings or omitting critical pathological observations, which are clinically unacceptable and remain the primary barrier to real-world deployment~\cite{bannur2024maira,wang2023r2gengpt,wang2024r2gencsr}.

Inspired by previous work~\cite{li2019knowledge,wang2022prior}, we hypothesize that this unreliability stems from coupling two distinct cognitive processes in a single model: visual feature recognition and medical language organization. We propose that by decoupling these tasks—first compelling the model to identify all salient clinical facts before composing the report—we can significantly enhance factual accuracy.

A primary obstacle is the lack of large-scale datasets pairing medical images with exhaustive key-finding labels. This is especially pronounced in disease-focused settings, where relevant categories are prohibitively expensive to annotate manually. Prior label-guided methods such as TieNet~\cite{wang2018tienet} assume a fixed vocabulary tightly coupled to specific datasets, limiting adaptability and rendering them incompatible with modern MLLM architectures.

To overcome this, we introduce Fact-Flow, a novel framework that improves the factual accuracy of MLLM-based report generation through multi-label guidance. The main contributions are as follows.
\begin{itemize}
    \item We propose Fact-Flow, which improves MLLM-based report generation via explicit multi-label clinical finding conditioning.
    \item We design a fully automated, LLM-bootstrapped data pipeline that constructs a large-scale (image, multi-label) dataset from existing image-report pairs without manual annotation.
    \item We validate Fact-Flow on two disease-focused datasets (ophthalmology and tuberculosis), demonstrating consistent improvements over state-of-the-art methods in NLG metrics and, where applicable, clinical efficacy.
\end{itemize}
\begin{figure}[t]
    \centering
    \includegraphics[width=\textwidth]{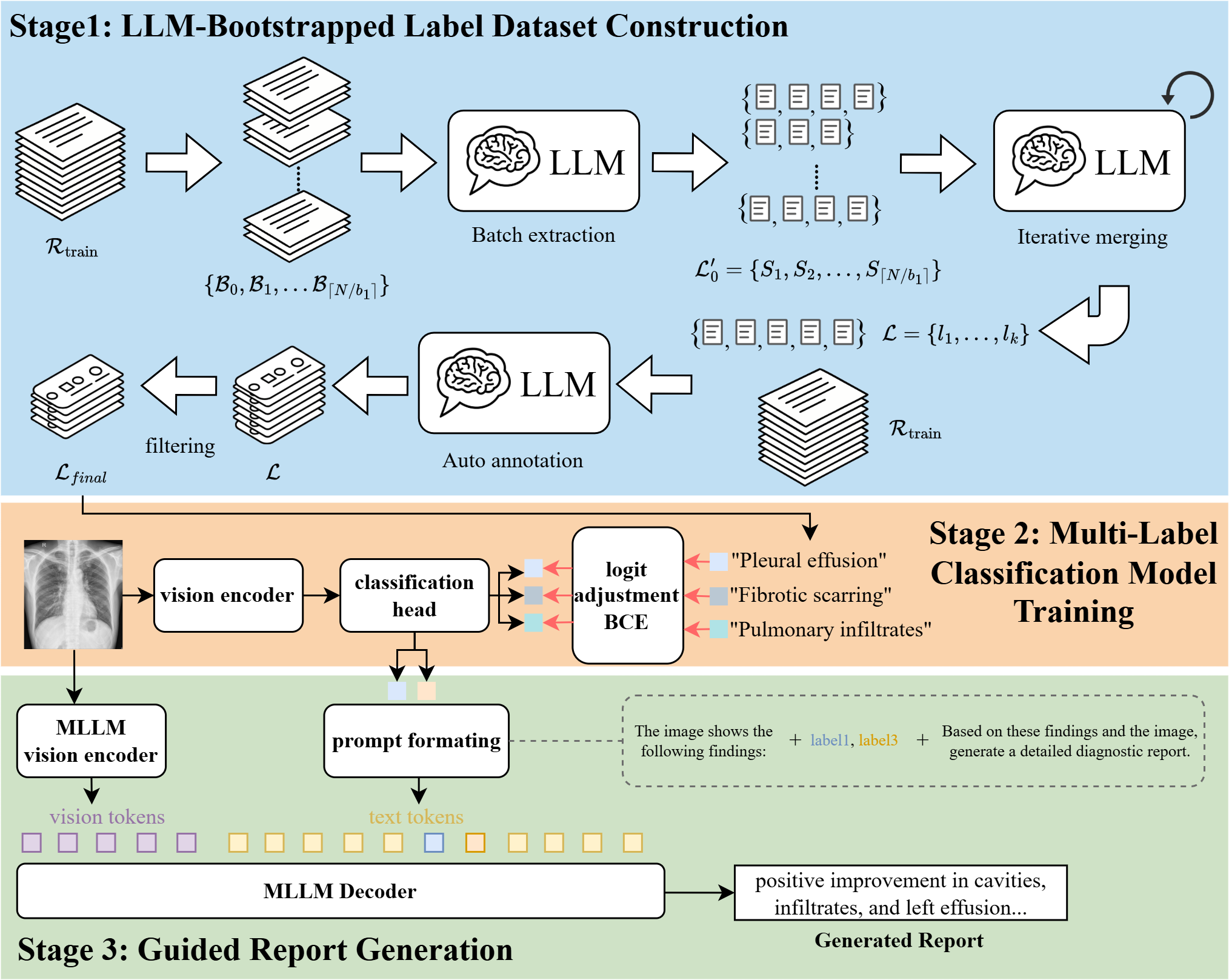}
    \caption{The overall framework of Fact-Flow}
    \label{fig:pipeline}
\end{figure}

\section{Method}
\label{sec:method}

We present \textbf{Fact-Flow}, a framework that enhances MLLM-based medical report generation through multi-label guidance. Given a training set of medical image--report pairs $\{(I_i, R_i)\}_{i=1}^N$, our goal is to generate a diagnostic report $\hat{R}$ for a new image $I$ that accurately captures all clinical findings. Rather than mapping $I \rightarrow R$ end-to-end, Fact-Flow introduces an intermediate multi-label representation $Y = \{y_1, \ldots, y_k\} \in \{0,1\}^k$, where each $y_j$ denotes the presence or absence of a specific clinical finding. As illustrated in Figure~\ref{fig:pipeline}, the framework comprises three stages: (1) LLM-bootstrapped label dataset construction, (2) multi-label classification model training, and (3) label-guided report generation, with the predicted labels serving as factual grounding to reduce hallucinations and improve finding recall.

\subsection{Stage 1: LLM-Bootstrapped Multi-Label Dataset Construction}

Since manually annotating large-scale reports with fine-grained clinical labels is prohibitively expensive, we construct the label dataset automatically via an LLM-driven pipeline (see Stage 1 in Figure~\ref{fig:pipeline}). The pipeline proceeds in two steps.

\textbf{Taxonomy extraction.} We first build a unified label taxonomy $\mathcal{L} = \{l_1, \ldots, l_k\}$ from the training reports. Processing all reports in a single LLM call is infeasible due to context-length constraints, so we partition $\{R_i\}_{i=1}^N$ into batches of size $b$ and prompt the LLM to extract clinically significant concepts---including diseases, pathological features, anatomical locations, and severity descriptors---from each batch independently. The resulting per-batch label sets inevitably contain synonyms and redundancies, so we consolidate them through iterative hierarchical merging: at each round, sets are grouped within a token budget $\kappa$ and the LLM is prompted to perform synonym normalization and deduplication, repeating until a single canonical taxonomy $\mathcal{L}$ is obtained.

\textbf{Report annotation and filtering.} Given $\mathcal{L}$, we annotate each training report $R_i$ by prompting the LLM to identify all labels explicitly or implicitly mentioned, producing a binary vector $Y_i \in \{0,1\}^k$ where $Y_i[j]=1$ if label $l_j$ is present. To reduce annotation noise and mitigate long-tail issues, we apply frequency-based filtering and retain only labels that appear in at least $\theta$ reports. This yields the final bootstrapped dataset $\mathcal{D}_{\text{MLC}} = \{(I_i, Y_i)\}_{i=1}^N$ used to train the guidance model in Stage 2.

\subsection{Stage 2: Guidance Model Training}

In this stage (see Stage 2 in Figure~\ref{fig:pipeline}), we train a multi-label classification model $f_{\text{MLC}}$, formally defined as $f_{\text{MLC}}(I) = \sigma(W_{\text{cls}} \cdot \phi(I))$, to predict clinical findings from medical images, where $\phi(\cdot)$ is a pre-trained vision encoder (DINOv3~\cite{simeoni2025dinov3} with a ConvNeXt backbone~\cite{liu2022convnet}) and $\sigma(\cdot)$ denotes the sigmoid function applied independently per label.

Medical datasets exhibit severe class imbalance, with rare but critical findings appearing in fewer than 1\% of cases. Standard binary cross-entropy tends to overlook these tail classes. To address this, we adapt the logit adjustment method~\cite{menon2020long} to multi-label binary classification, shifting each raw logit $z_j$ by the log-odds of its empirical frequency $p_j$ before computing the loss:
\begin{equation}
\tilde{z}_j = z_j + \tau\log\tfrac{p_j}{1-p_j}
\end{equation}
where $p_j = \frac{1}{N}\sum_{i=1}^{N}Y_i[j]$ is the empirical label frequency and $\tau$ is a temperature parameter (set to $1$). The adjusted logits $\tilde{z}_j$ are then used in the standard binary cross-entropy loss, re-balancing the decision boundary to improve both precision and recall on tail labels.

\subsection{Stage 3: Guided Report Generation}

In the final stage (see Stage 3 in Figure~\ref{fig:pipeline}), we fine-tune a multimodal large language model (MLLM) to generate diagnostic reports conditioned on both visual features and predicted clinical findings.

During training, the bootstrapped ground-truth labels $Y_i$ are serialized into a natural-language prompt (e.g., \textit{``The image shows the following findings: [label$_a$], [label$_b$], [label$_c$]. Based on these findings and the image, generate a detailed diagnostic report.''}), which is prepended to the generation target. The model is optimized with the standard next-token prediction objective, learning to condition its output on both the visual input and the explicit label guidance.

At inference time, the ground-truth labels are unavailable. Instead, the predicted labels $\hat{Y} = f_{\text{MLC}}(I)$ from Stage 2 are serialized into the same prompt format and used to guide generation, grounding the report in explicitly identified factual findings.
\section{Experiments}
\label{sec:experiments}

\subsection{Dataset and Evaluation Metrics}
We evaluate Fact-Flow on two medical imaging datasets. The first is the publicly available tuberculosis chest X-ray dataset~\cite{jaeger2014two}, comprising 561/80/160 frontal chest X-rays for training, validation, and testing, each paired with a diagnostic report describing tuberculosis-related findings. The second is a multimodal ophthalmology dataset collected from a clinical institution, consisting of 1,854/206/515 cases for training, validation, and testing, with IRB approval for research use. Each case provides three imaging modalities—fundus photography, Optical Coherence Tomography (OCT), and Optical Coherence Tomography Angiography (OCTA)—paired with a detailed Chinese-language diagnostic report authored by ophthalmologists describing myopia-related complications and retinal changes.

We assess report quality using two sets of metrics. For natural language generation (NLG), we employ BLEU-1 through BLEU-4~\cite{papineni2002bleu}, ROUGE-L~\cite{lin2004rouge}, CIDEr~\cite{vedantam2015cider}, and METEOR~\cite{banerjee2005meteor}. For clinical efficacy on the tuberculosis dataset, we adopt RadFact~\cite{bannur2024maira}, which leverages large language models to extract clinical entities from both generated and reference reports and computes Precision, Recall, and F1-score without requiring predefined disease-specific categories. No established clinical efficacy metric is currently available for Chinese ophthalmology reports, so only NLG metrics are reported for that dataset.

\subsection{Implementation Details}
For Stage 1 label bootstrapping, we use GPT-5-mini to extract and merge clinical findings from training reports with $b=200$, $\kappa=200$, and a merging threshold $\theta=15$, yielding 7 labels for tuberculosis and 42 for ophthalmology datasets. For Stage 2, the multi-label classifier $f_{\text{MLC}}$ employs DINOv3-ConvNeXt-Base~\cite{simeoni2025dinov3,liu2022convnet} as the visual encoder with full fine-tuning and logit adjustment~\cite{menon2020long} ($\tau=1.0$) for long-tail optimization, trained with a learning rate of $2\times10^{-5}$. For the ophthalmology dataset, we use separate encoders for each modality (fundus, OCT, OCTA) with feature fusion. For Stage 3, we fine-tune three MLLMs—Qwen2.5-VL 7B~\cite{bai2025qwen2}, MedGemma 4B~\cite{sellergren2025medgemma}, and LLaVA-Med-v1.5-Mistral 7B~\cite{li2023llava}—using LoRA~\cite{hu2022lora} on the decoder, vision encoder, and projector; all models are trained with the AdamW optimizer (learning rate $2\times10^{-4}$, weight decay $0.1$) using BFloat16 mixed precision on four NVIDIA RTX A6000 GPUs.

\begin{table}[t]
    \centering
    \caption{Comparison with state-of-the-art methods on the Tuberculosis dataset. ``+FF'' denotes our Fact-Flow approach.}
    \label{tab:tb_comparison}
    \resizebox{\linewidth}{!}{%
    \begin{tabular}{l|l|ccc|cccc}
    \toprule
    \multirow{2}{*}{\textbf{Type}} & \multirow{2}{*}{\textbf{Method}} & \multicolumn{3}{c|}{\textbf{Clinical Efficacy}} & \multicolumn{4}{c}{\textbf{Natural Language Generation}} \\
    \cline{3-9}
    & & \textbf{F1} & \textbf{Prec.} & \textbf{Recall} & \textbf{B-4} & \textbf{R-L} & \textbf{CIDEr} & \textbf{METEOR} \\
    \midrule
    \multirow{5}{*}{\textit{Traditional}} 
    & R2Gen & 0.1909 & 0.2273 & 0.1646 & 0.0682 & 0.6423 & 1.7061 & 0.2131 \\
    & R2GPT & 0.2105 & 0.1310 & \underline{0.5361} & 0.1032 & 0.3101 & 1.1659 & 0.1813 \\
    & CvT2DistilGPT2 & 0.2128 & 0.2333 & 0.1957 & 0.1655 & 0.5752 & 1.7217 & 0.1896 \\
    & R2GenCSR & 0.2041 & \underline{0.3000} & 0.1546 & 0.1478 & 0.6333 & 1.8396 & 0.1953 \\
    & R2GenCMN & 0.2334 & 0.2903 & 0.1952 & 0.0603 & \underline{0.6500} & 1.7216 & 0.1990 \\
    \midrule
    \multirow{3}{*}{\textit{MLLM}} 
    & MedGemma & 0.2266 & 0.2422 & 0.2130 & 0.0795 & \textbf{0.6566} & 1.7307 & 0.1759 \\
    & LLaVA-Med & 0.0000 & 0.0000 & 0.0000 & 0.0242 & 0.4691 & 1.1719 & 0.1177 \\
    & Qwen2.5-VL & 0.0286 & 1.0000 & 0.0145 & 0.0265 & 0.4779 & 1.2035 & 0.1141 \\
    \midrule
    \multirow{2}{*}{\textit{Zero-Shot}}
    & Gemini-2.5-Flash & 0.0677 & 0.0525 & 0.0952 & 0.0014 & 0.0383 & 0.0102 & 0.0295 \\
    & Gemini-2.5-Pro & 0.0713 & 0.0457 & 0.1621 & 0.0008 & 0.0301 & 0.0003 & 0.0245 \\
    \midrule
    \multirow{3}{*}{\textit{\textbf{Fact-Flow}}} 
    & MedGemma + FF & \textbf{0.3055} & 0.3070 & \textbf{0.3041} & \textbf{0.2290} & 0.6243 & \textbf{2.0664} & \textbf{0.2305} \\
    & LLaVA-Med + FF & 0.2016 & 0.1974 & 0.2059 & 0.1287 & 0.6018 & 1.5869 & 0.1869 \\
    & Qwen2.5-VL + FF & \underline{0.2831} & \textbf{0.3535} & 0.2361 & \underline{0.2129} & 0.6107 & \underline{1.9434} & \underline{0.2236} \\
    \bottomrule
    \end{tabular}
    }
    \end{table}

    \begin{table}[t]
        \centering
        \caption{Comparison with state-of-the-art methods on the Ophthalmology dataset. ``+FF'' denotes our Fact-Flow approach.}
        \label{tab:opht_comparison}
        \resizebox{0.85\linewidth}{!}{%
        \begin{tabular}{lcccccc}
        \toprule
        \textbf{Method} & \textbf{BLEU-1} & \textbf{BLEU-2} & \textbf{BLEU-3} & \textbf{BLEU-4} & \textbf{ROUGE-L} & \textbf{CIDEr} \\
        \midrule
        \multicolumn{7}{l}{\textit{Traditional Methods}} \\
        R2Gen & 0.6004 & 0.4703 & 0.3807 & 0.3138 & 0.3605 & 0.0232 \\
        R2GenCMN & 0.6232 & 0.5178 & 0.4420 & 0.3838 & 0.7065 & 0.0650 \\
        R2GPT & 0.7324 & 0.6446 & 0.5777 & 0.5231 & 0.8389 & \underline{0.2589} \\
        CvT2DistilGPT2 & 0.6249 & 0.5224 & 0.4485 & 0.3904 & 0.4131 & 0.0268 \\
        R2GenCSR & 0.7216 & 0.6372 & 0.5710 & 0.5165 & 0.8386 & 0.1786 \\
        \midrule
        \multicolumn{7}{l}{\textit{MLLM}} \\
        MedGemma & 0.6489 & 0.5562 & 0.4867 & 0.4318 & 0.7921 & 0.0986 \\
        LLaVA-Med & 0.7242 & 0.6266 & 0.5521 & 0.4935 & 0.8345 & 0.1741 \\
        Qwen2.5-VL & 0.6873 & 0.5853 & 0.5098 & 0.4506 & 0.7947 & 0.1382 \\
        \midrule
        \multicolumn{7}{l}{\textit{\textbf{Fact-Flow}}} \\
        MedGemma + FF & 0.7079 & 0.6267 & 0.5627 & 0.5099 & 0.8272 & 0.2306 \\
        LLaVA-Med + FF & \textbf{0.7685} & \underline{0.6798} & \underline{0.6103} & \underline{0.5533} & \underline{0.8457} & 0.2350 \\
        Qwen2.5-VL + FF & \underline{0.7681} & \textbf{0.6820} & \textbf{0.6133} & \textbf{0.5567} & \textbf{0.8558} & \textbf{0.2759} \\
        \bottomrule
        \end{tabular}
        }
        \end{table}
    
\begin{table}[t]
    \centering
    \caption{Case study of generated reports on the Tuberculosis dataset. Text with green background indicates correct findings matching ground truth, while red background indicates incorrect or missing findings.}
    \label{tab:case_study}
    \setlength{\tabcolsep}{4pt}
    \resizebox{\linewidth}{!}{%
    \begin{tabular}{l|p{4cm}|p{4cm}|p{3cm}|p{2cm}}
    \toprule
    \textbf{Chest X-ray} & \textbf{Ground Truth} & \textbf{MLC Prediction} & \textbf{MedGemma + Fact-Flow} & \textbf{MedGemma} \\
    \midrule
    \adjustbox{valign=t}{\includegraphics[width=2.5cm]{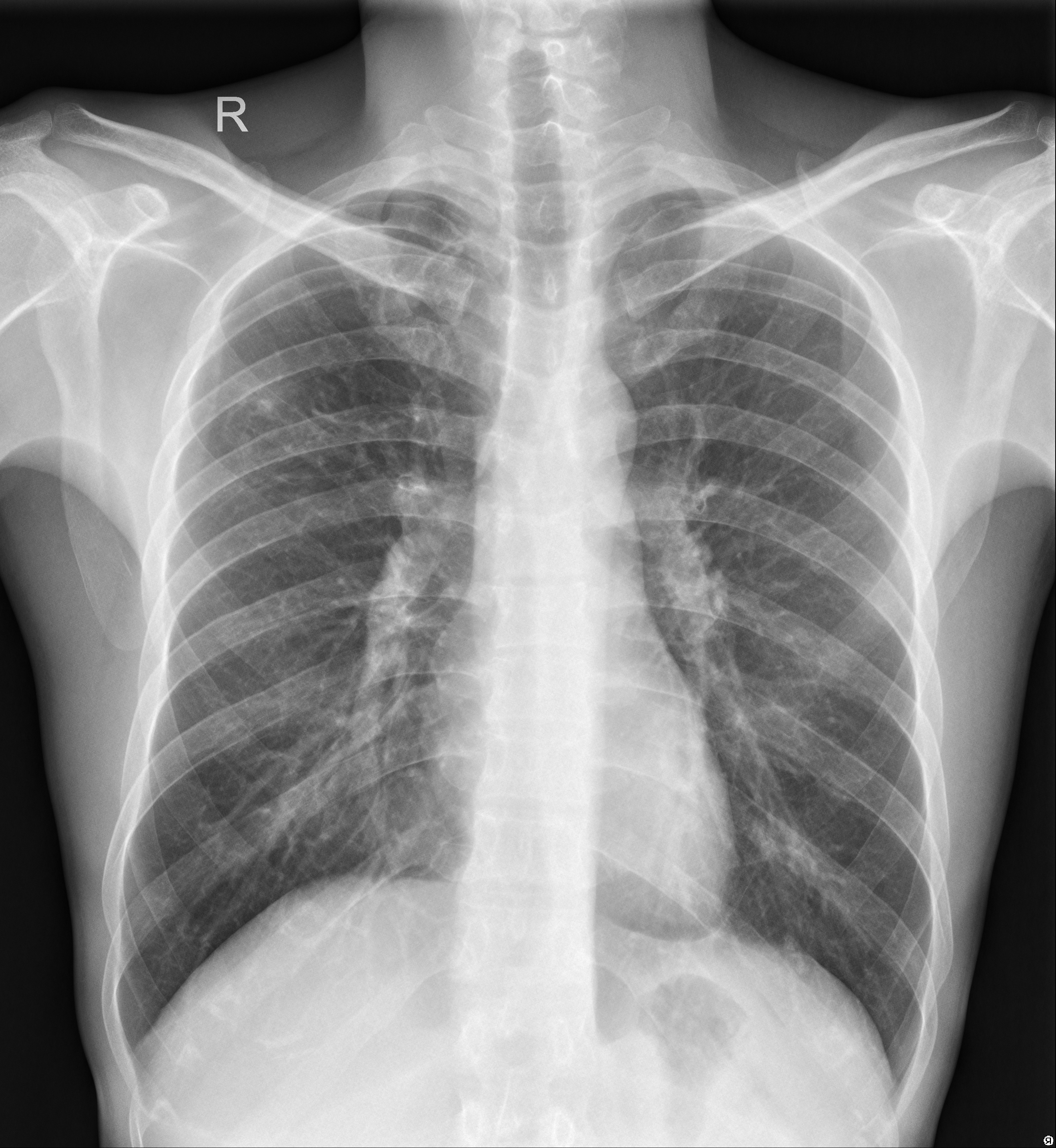}} & 
    \hlblue{Right secondary PTB in the upper and middle fields} & 
    \hlgreen{Pulmonary tuberculosis}, \hlgreen{Secondary (post-primary) pulmonary tuberculosis} & 
    \hlgreen{secondary PTB} in the \hlgreen{right upper field} & 
    \hlred{bilateral} \hlgreen{PTB} \\
    \cmidrule(lr){1-5}
    RadFact F1 & & & \textbf{1} & 0 \\
    \midrule
    \adjustbox{valign=t}{\includegraphics[width=2.5cm]{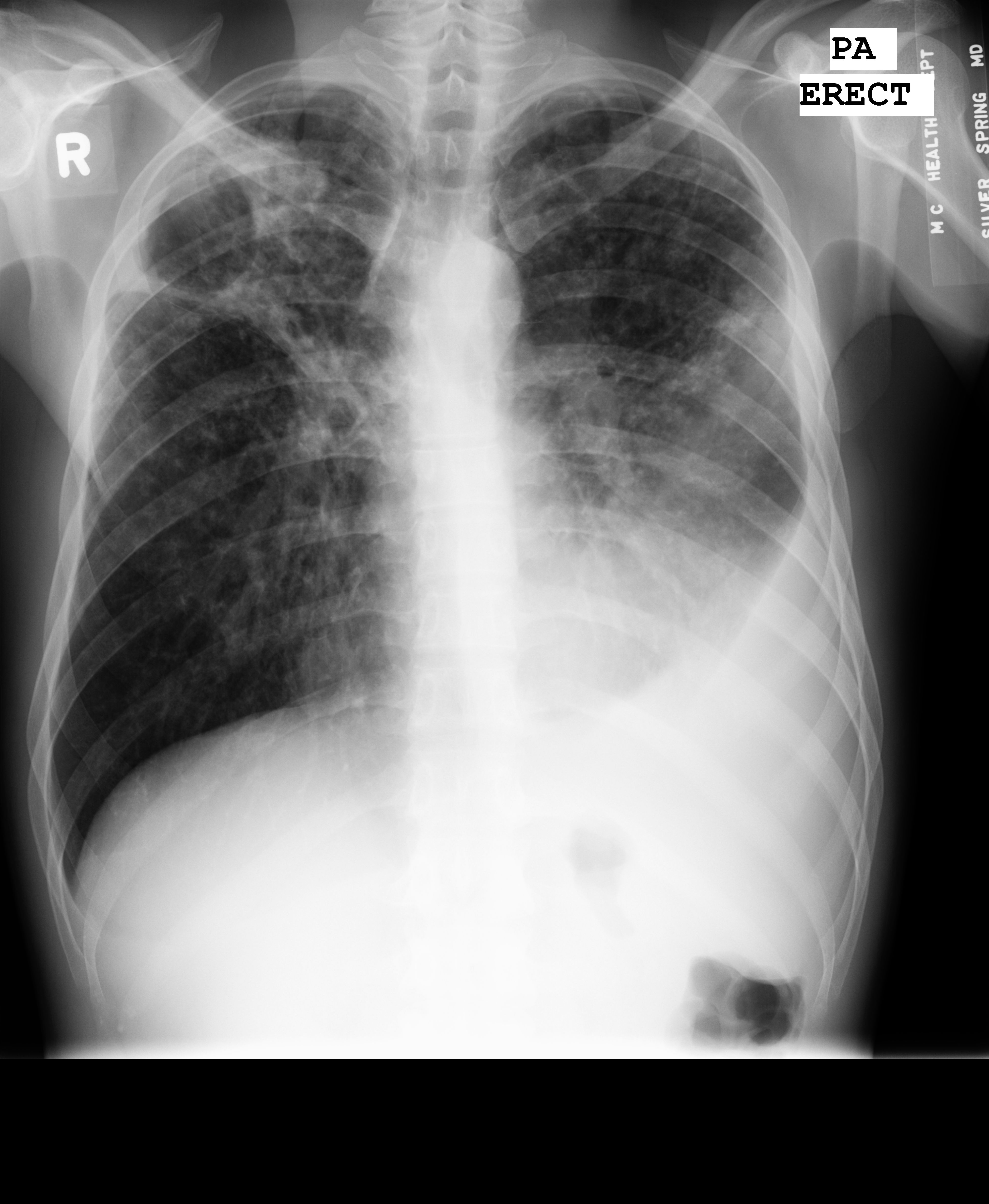}} & 
    extensive \hlblue{infiltrates bilaterally} with large \hlblue{cavity} in RUL and a moderate \hlblue{pleural effusion on the left}. AFB smears and RNA probes pos for MTB. \hlblue{Active TB}, \hlblue{cavitary}. & 
    \hlgreen{Pulmonary tuberculosis}, \hlgreen{Pulmonary cavitation}, \hlgreen{Pulmonary infiltrates}, \hlgreen{Pleural effusion} & 
    \hlred{bilateral} \hlgreen{PTB} with \hlgreen{cavitation} and left \hlgreen{pleural effusion} & 
    \hlred{bilateral} \hlgreen{PTB} \\
    \cmidrule(lr){1-5}
    RadFact F1 & & & \textbf{0.67} & 0 \\
    \bottomrule
    \end{tabular}
    }
\end{table}

\subsection{Comparison with State-of-the-Art Methods}
We compare Fact-Flow against traditional MRG models (R2Gen~\cite{chen2020generating}, R2GenCMN~\cite{chen2021cross}, R2GPT~\cite{wang2023r2gengpt}, R2GenCSR~\cite{wang2024r2gencsr}, CvT2DistilGPT2~\cite{nicolson2023improving}), MLLM baselines (Qwen2.5-VL, MedGemma, LLaVA-Med fine-tuned directly on image-report pairs without factual guidance), and powerful closed-source VLMs (Gemini-2.5-Flash and Gemini-2.5-Pro) evaluated in a zero-shot setting. On the tuberculosis dataset (Table~\ref{tab:tb_comparison}), our approach consistently improves all three MLLMs across NLG and clinical efficacy metrics. \textbf{MedGemma + Fact-Flow} achieves the best overall performance, while vanilla MLLM baselines suffer from mode collapse—Qwen2.5-VL yields perfect precision but near-zero recall, and LLaVA-Med produces zero clinical efficacy scores—a problem our factual guidance directly addresses. Despite their strong general capabilities, Gemini-2.5-Flash and Gemini-2.5-Pro perform poorly in the zero-shot setting, underscoring the necessity of domain-specific training for medical report generation. On the ophthalmology dataset (Table~\ref{tab:opht_comparison}), \textbf{Qwen2.5-VL + Fact-Flow} achieves the best results on most NLG metrics, demonstrating the effectiveness of our approach across complex multimodal scenarios. Qualitative examples in Table~\ref{tab:case_study} further confirm that factual guidance enables more precise localization of findings (e.g., disease laterality, anatomical region) compared to the vanilla MLLM baseline.

\begin{table}[t]
    \centering
    \caption{Analysis of visual and factual guidance on the Tuberculosis dataset using Qwen2.5-VL. ``Pred'' and ``GT'' denote predicted and ground-truth labels, respectively.}
    \label{tab:ablation}
    \resizebox{\linewidth}{!}{%
    \begin{tabular}{cc|ccc|cccccc}
    \toprule
    \multicolumn{2}{c|}{\textbf{Input Modality}} & \multicolumn{3}{c|}{\textbf{Clinical Efficacy}} & \multicolumn{6}{c}{\textbf{Natural Language Generation}} \\
    \cline{1-11}
    \textbf{Image} & \textbf{Label} & \textbf{F1} & \textbf{Prec.} & \textbf{Recall} & \textbf{B-2} & \textbf{B-3} & \textbf{B-4} & \textbf{R-L} & \textbf{CIDEr} & \textbf{METEOR} \\
    \midrule
    \multicolumn{11}{l}{\textit{Practical Configurations}} \\
    \checkmark & -- & 0.0286 & 1.0000 & 0.0145 & 0.0265 & 0.0265 & 0.0265 & 0.4779 & 1.2035 & 0.1141 \\
    -- & Pred & 0.2115 & 0.2656 & 0.1757 & 0.2895 & 0.2409 & 0.2074 & 0.5943 & 1.9209 & 0.2188 \\
    \checkmark & Pred & \textbf{0.2831} & \textbf{0.3535} & \textbf{0.2361} & \textbf{0.2951} & \textbf{0.2455} & \textbf{0.2129} & \textbf{0.6107} & \textbf{1.9434} & \textbf{0.2236} \\
    \midrule
    \multicolumn{11}{l}{\textit{Oracle Upper Bounds}} \\
    -- & GT & \textit{0.3750} & \textit{0.5000} & \textit{0.3000} & \textit{0.3658} & \textit{0.3026} & \textit{0.2618} & \textit{0.6990} & \textit{2.3198} & \textit{0.2691} \\
    \checkmark & GT & \textit{0.4421} & \textit{0.5116} & \textit{0.3892} & \textit{0.3727} & \textit{0.3095} & \textit{0.2680} & \textit{0.7198} & \textit{2.4420} & \textit{0.2746} \\
    \bottomrule
    \end{tabular}
    }
    \end{table}

\subsection{Analysis of Visual and Factual Guidance}
We evaluate five configurations on the tuberculosis dataset using Qwen2.5-VL (Table~\ref{tab:ablation}), varying input modalities (image and/or label) and label sources (predicted vs.\ ground-truth). The \textbf{Image Only} baseline suffers from mode collapse despite high precision, as MLLMs generate overly conservative reports without factual grounding. Introducing predicted labels (\textbf{Label Only (Pred)}) substantially improves both clinical efficacy and NLG metrics, and combining image with predicted labels (\textbf{Image + Label (Pred)})—our full Fact-Flow approach—achieves the best practical performance, confirming that visual context and factual guidance are complementary. The performance gap between predicted and ground-truth label conditions highlights label quality as the key bottleneck, and the oracle \textbf{Image + Label (GT)} consistently outperforms \textbf{Label Only (GT)}, indicating that visual information provides spatial details that discrete labels alone cannot capture.

\subsection{Intermediate Stage Analysis}

\begin{table}[t]
\centering
\caption{Intermediate stage analysis. \textbf{Left}: Intrinsic quality evaluation of Stage 1 on the ophthalmology dataset (all metrics verified by a professional ophthalmologist). Report-level metrics (Coverage, LLM-Match Accuracy) are measured on 50 randomly sampled reports; label-level metrics (Redundancy, Clinical Validity) characterize $\mathcal{L}$ itself. \textbf{Right}: Stage 2 MLC classification performance on test sets.}
\label{tab:stage_analysis}
\small
\begin{minipage}[t]{0.5\linewidth}
\centering
\begin{tabular}{lr}
\hline
\textbf{Stage 1} & \textbf{Result} \\
\hline
\multicolumn{2}{l}{\textit{Report-level (50 sampled reports)}} \\
\quad Coverage & 100\% (50/50) \\
\quad LLM-Match Accuracy & 100\% (50/50) \\
\hline
\multicolumn{2}{l}{\textit{Label-level ($\mathcal{L}$)}} \\
\quad Redundancy & 7.5\% \\
\quad Valid \& decidable & 80\% \\
\quad Valid, context-dependent & 15\% \\
\quad Overly generic & 5\% \\
\quad Invalid / pseudo-concept & 0\% \\
\hline
\end{tabular}
\end{minipage}%
\begin{minipage}[t]{0.42\linewidth}
\centering
\begin{tabular}{lcc}
\hline
\textbf{Stage 2} & \textbf{Macro-F1} & \textbf{Micro-F1} \\
\hline
Tuberculosis    & 0.5227 & 0.7328 \\
Ophthalmology   & 0.7071 & 0.8426 \\
\hline
\end{tabular}
\end{minipage}
\end{table}

We validate each intermediate pipeline stage (Table~\ref{tab:stage_analysis}). \textbf{Stage 1:} All metrics are verified by a professional ophthalmologist on the ophthalmology dataset. (1) \textbf{Coverage} measures whether each report's key clinical information can be expressed by a subset of labels in $\mathcal{L}$, rated as \emph{fully} / \emph{partially} / \emph{not covered} on 50 randomly sampled reports; (2) \textbf{LLM-Match Accuracy} checks whether LLM-Match assigns correct labels (set-level) on the same 50 reports; (3) \textbf{Redundancy} estimates the proportion of clinically synonymous label pairs within $\mathcal{L}$; and (4) \textbf{Clinical Validity} categorizes each label as \emph{valid \& decidable}, \emph{valid but context-dependent}, \emph{overly generic}, or \emph{invalid/pseudo-concept}. Results confirm 100\% coverage and matching accuracy, low redundancy (7.5\%), and 80\% of labels valid and decidable with none invalid, validating the bootstrapped vocabulary. \textbf{Stage 2:} The MLC achieves Macro-/Micro-F1 of 0.52/0.73 on tuberculosis and 0.71/0.84 on ophthalmology, providing sufficient grounding for Stage 3.
\section{Conclusion}
\label{sec:conclusion}

To address the critical challenge of factual instability in Multimodal Large Language Models (MLLMs) when applied to medical report generation, we propose Fact-Flow, a novel framework that enhances factual accuracy by decoupling visual feature recognition from textual composition. Our core contributions include an LLM-bootstrapped pipeline for automatic multi-label dataset creation and a multi-label classification model that provides an explicit ``Fact-Flow'' to guide the MLLM. Fact-Flow is a plug-and-play framework compatible with any MLLM architecture, particularly suited to clinical scenarios where reports revolve around targeted and enumerable finding categories. Experiments on two disease-focused datasets with three MLLMs (Qwen2.5-VL, MedGemma, and LLaVA-Med) demonstrate that Fact-Flow consistently and significantly improves the factual correctness of generated reports over state-of-the-art methods and direct fine-tuning, without compromising textual quality.

\bibliographystyle{splncs04}
\bibliography{main.bib}

@inproceedings{jing2018automatic,
  title={On the automatic generation of medical imaging reports},
  author={Jing, Baoyu and Xie, Pengtao and Xing, Eric},
  booktitle={Proceedings of the 56th annual meeting of the association for computational linguistics (volume 1: long papers)},
  pages={2577--2586},
  year={2018}
}

@inproceedings{wang2018tienet,
  title={Tienet: Text-image embedding network for common thorax disease classification and reporting in chest x-rays},
  author={Wang, Xiaosong and Peng, Yifan and Lu, Le and Lu, Zhiyong and Summers, Ronald M},
  booktitle={Proceedings of the IEEE conference on computer vision and pattern recognition},
  pages={9049--9058},
  year={2018}
}

@inproceedings{xue2018multimodal,
  title={Multimodal recurrent model with attention for automated radiology report generation},
  author={Xue, Yuan and Xu, Tao and Rodney Long, L and Xue, Zhiyun and Antani, Sameer and Thoma, George R and Huang, Xiaolei},
  booktitle={International Conference on Medical Image Computing and Computer-Assisted Intervention},
  pages={457--466},
  year={2018},
  organization={Springer}
}

@article{chen2020generating,
  title={Generating radiology reports via memory-driven transformer},
  author={Chen, Zhihong and Song, Yan and Chang, Tsung-Hui and Wan, Xiang},
  journal={arXiv preprint arXiv:2010.16056},
  year={2020}
}

@inproceedings{chen2021cross,
  title={Cross-modal memory networks for radiology report generation},
  author={Chen, Zhihong and Shen, Yaling and Song, Yan and Wan, Xiang},
  booktitle={Proceedings of the 59th annual meeting of the association for computational linguistics and the 11th international joint conference on natural language processing (volume 1: long papers)},
  pages={5904--5914},
  year={2021}
}

@article{nooralahzadeh2021progressive,
  title={Progressive transformer-based generation of radiology reports},
  author={Nooralahzadeh, Farhad and Gonzalez, Nicolas Perez and Frauenfelder, Thomas and Fujimoto, Koji and Krauthammer, Michael},
  journal={arXiv preprint arXiv:2102.09777},
  year={2021}
}

@inproceedings{liu2021exploring,
  title={Exploring and distilling posterior and prior knowledge for radiology report generation},
  author={Liu, Fenglin and Wu, Xian and Ge, Shen and Fan, Wei and Zou, Yuexian},
  booktitle={Proceedings of the IEEE/CVF conference on computer vision and pattern recognition},
  pages={13753--13762},
  year={2021}
}

@article{wang2023r2gengpt,
  title={R2gengpt: Radiology report generation with frozen llms},
  author={Wang, Zhanyu and Liu, Lingqiao and Wang, Lei and Zhou, Luping},
  journal={Meta-Radiology},
  volume={1},
  number={3},
  pages={100033},
  year={2023},
  publisher={Elsevier}
}

@article{wang2024r2gencsr,
  title={R2gencsr: Retrieving context samples for large language model based x-ray medical report generation},
  author={Wang, Xiao and Li, Yuehang and Wang, Fuling and Wang, Shiao and Li, Chuanfu and Jiang, Bo},
  journal={arXiv preprint arXiv:2408.09743},
  year={2024}
}

@article{nicolson2023improving,
  title={Improving chest X-ray report generation by leveraging warm starting},
  author={Nicolson, Aaron and Dowling, Jason and Koopman, Bevan},
  journal={Artificial intelligence in medicine},
  volume={144},
  pages={102633},
  year={2023},
  publisher={Elsevier}
}

@article{bai2025qwen2,
  title={Qwen2. 5-vl technical report},
  author={Bai, Shuai and Chen, Keqin and Liu, Xuejing and Wang, Jialin and Ge, Wenbin and Song, Sibo and Dang, Kai and Wang, Peng and Wang, Shijie and Tang, Jun and others},
  journal={arXiv preprint arXiv:2502.13923},
  year={2025}
}

@article{li2023llava,
  title={Llava-med: Training a large language-and-vision assistant for biomedicine in one day},
  author={Li, Chunyuan and Wong, Cliff and Zhang, Sheng and Usuyama, Naoto and Liu, Haotian and Yang, Jianwei and Naumann, Tristan and Poon, Hoifung and Gao, Jianfeng},
  journal={Advances in Neural Information Processing Systems},
  volume={36},
  pages={28541--28564},
  year={2023}
}

@article{sellergren2025medgemma,
  title={Medgemma technical report},
  author={Sellergren, Andrew and Kazemzadeh, Sahar and Jaroensri, Tiam and Kiraly, Atilla and Traverse, Madeleine and Kohlberger, Timo and Xu, Shawn and Jamil, Fayaz and Hughes, C{\'\i}an and Lau, Charles and others},
  journal={arXiv preprint arXiv:2507.05201},
  year={2025}
}

@inproceedings{papineni2002bleu,
  title={Bleu: a method for automatic evaluation of machine translation},
  author={Papineni, Kishore and Roukos, Salim and Ward, Todd and Zhu, Wei-Jing},
  booktitle={Proceedings of the 40th annual meeting of the Association for Computational Linguistics},
  pages={311--318},
  year={2002}
}

@inproceedings{lin2004rouge,
  title={Rouge: A package for automatic evaluation of summaries},
  author={Lin, Chin-Yew},
  booktitle={Text summarization branches out},
  pages={74--81},
  year={2004}
}

@inproceedings{vedantam2015cider,
  title={Cider: Consensus-based image description evaluation},
  author={Vedantam, Ramakrishna and Lawrence Zitnick, C and Parikh, Devi},
  booktitle={Proceedings of the IEEE conference on computer vision and pattern recognition},
  pages={4566--4575},
  year={2015}
}

@inproceedings{banerjee2005meteor,
  title={METEOR: An automatic metric for MT evaluation with improved correlation with human judgments},
  author={Banerjee, Satanjeev and Lavie, Alon},
  booktitle={Proceedings of the acl workshop on intrinsic and extrinsic evaluation measures for machine translation and/or summarization},
  pages={65--72},
  year={2005}
}

@article{bannur2024maira,
  title={Maira-2: Grounded radiology report generation},
  author={Bannur, Shruthi and Bouzid, Kenza and Castro, Daniel C and Schwaighofer, Anton and Thieme, Anja and Bond-Taylor, Sam and Ilse, Maximilian and P{\'e}rez-Garc{\'\i}a, Fernando and Salvatelli, Valentina and Sharma, Harshita and others},
  journal={arXiv preprint arXiv:2406.04449},
  year={2024}
}

@article{jaeger2014two,
  title={Two public chest X-ray datasets for computer-aided screening of pulmonary diseases},
  author={Jaeger, Stefan and Candemir, Sema and Antani, Sameer and W{\'a}ng, Y{\`\i}-Xi{\'a}ng J and Lu, Pu-Xuan and Thoma, George},
  journal={Quantitative imaging in medicine and surgery},
  volume={4},
  number={6},
  pages={475},
  year={2014}
}

@inproceedings{miura2021improving,
  title={Improving factual completeness and consistency of image-to-text radiology report generation},
  author={Miura, Yasuhide and Zhang, Yuhao and Tsai, Emily and Langlotz, Curtis and Jurafsky, Dan},
  booktitle={Proceedings of the 2021 Conference of the North American Chapter of the Association for Computational Linguistics: Human Language Technologies},
  pages={5288--5304},
  year={2021}
}

@article{liu2023visual,
  title={Visual instruction tuning},
  author={Liu, Haotian and Li, Chunyuan and Wu, Qingyang and Lee, Yong Jae},
  journal={Advances in neural information processing systems},
  volume={36},
  pages={34892--34916},
  year={2023}
}

@article{wang2022prior,
  title={Prior knowledge enhances radiology report generation},
  author={Wang, Song and Tang, Liyan and Lin, Mingquan and Shih, George and Ding, Ying and Peng, Yifan},
  journal={AMIA Summits on Translational Science Proceedings},
  volume={2022},
  pages={486},
  year={2022}
}

@inproceedings{li2019knowledge,
  title={Knowledge-driven encode, retrieve, paraphrase for medical image report generation},
  author={Li, Christy Y and Liang, Xiaodan and Hu, Zhiting and Xing, Eric P},
  booktitle={Proceedings of the AAAI conference on artificial intelligence},
  volume={33},
  pages={6666--6673},
  year={2019}
}

@article{simeoni2025dinov3,
  title={Dinov3},
  author={Sim{\'e}oni, Oriane and Vo, Huy V and Seitzer, Maximilian and Baldassarre, Federico and Oquab, Maxime and Jose, Cijo and Khalidov, Vasil and Szafraniec, Marc and Yi, Seungeun and Ramamonjisoa, Micha{\"e}l and others},
  journal={arXiv preprint arXiv:2508.10104},
  year={2025}
}

@inproceedings{liu2022convnet,
  title={A convnet for the 2020s},
  author={Liu, Zhuang and Mao, Hanzi and Wu, Chao-Yuan and Feichtenhofer, Christoph and Darrell, Trevor and Xie, Saining},
  booktitle={Proceedings of the IEEE/CVF conference on computer vision and pattern recognition},
  pages={11976--11986},
  year={2022}
}

@article{menon2020long,
  title={Long-tail learning via logit adjustment},
  author={Menon, Aditya Krishna and Jayasumana, Sadeep and Rawat, Ankit Singh and Jain, Himanshu and Veit, Andreas and Kumar, Sanjiv},
  journal={arXiv preprint arXiv:2007.07314},
  year={2020}
}

@inproceedings{hu2022lora,
  title={LoRA: Low-Rank Adaptation of Large Language Models},
  author={Hu, Edward J and Shen, Yelong and Wallis, Phillip and Allen-Zhu, Zeyuan and Li, Yuanzhi and Wang, Shean and Wang, Lu and Chen, Weizhu},
  booktitle={International Conference on Learning Representations},
  year={2022}
}

\end{document}